\PassOptionsToPackage{table,xcdraw}{xcolor}
\documentclass[manuscript]{acmart}

\AtBeginDocument{%
  }

\usepackage{scalerel,graphicx,xparse}

\usepackage{amsmath}
\usepackage{cancel}
\usepackage[normalem]{ulem}
\usepackage{multirow}
\usepackage{makecell}
\usepackage{booktabs}
\usepackage[most]{tcolorbox}
\usepackage{array}
\usepackage{subcaption}
\usepackage{tabularx}
\usepackage{lscape}
\usepackage{rotating}

\newcolumntype{L}[1]{>{\raggedhright\let\newline\\\arraybackslash\hspace{0pt}}m{#1}}
\newcolumntype{C}[1]{>{\centering\let\newline\\\arraybackslash\hspace{0pt}}m{#1}}
\newcolumntype{R}[1]{>{\raggedleft\let\newline\\\arraybackslash\hspace{0pt}}m{#1}}

\usepackage{xcolor}
\begin{document}

\title{Right, No Matter Why: AI Fact-checking and AI Authority in Health-related Inquiry Settings}




\author{Elena Sergeeva}
\authornote{Both authors contributed equally to this research.}
\affiliation{%
  \institution{Massachusetts Institute of Technology}
  \streetaddress{77 Massachusetts Avenue}
  \city{Cambridge}
  \country{USA}}
\email{elenaser@mit.edu}

\author{Anastasia Sergeeva}
\authornotemark[1]
\affiliation{%
  \institution{University of Luxembourg}
  \streetaddress{2 Av. de l'Universite}
  \city{Esch-sur-Alzette}
  \country{Luxembourg}}
\email{anastasia.sergeeva@uni.lu}

\author{Huiyun Tang}
\affiliation{%
  \institution{University of Luxembourg}
  \streetaddress{2 Av. de l'Universite}
  \city{Esch-sur-Alzette}
  \country{Luxembourg}}
\email{huiyun.tang@uni.lu}

\author{Kerstin Bongard-Blanchy}
\affiliation{%
  \institution{Luxembourg Media and Digital Design Center}
  \streetaddress{4, rue Samuel Beckett}
  \city{Belvaux}
    \country{Luxembourg}}
\email{kerstin.bongard-blanchy@uni.lu}

\author{Peter Szolovits}
\affiliation{%
  \institution{Massachusetts Institute of Technology}
  \streetaddress{77 Massachusetts Avenue}
  \city{Cambridge}
  \country{USA}}
\email{psz@mit.edu}

\begin{abstract}
 Previous research on expert advice-taking shows that humans exhibit two contradictory behaviors: on the one hand, people tend to overvalue their own opinions undervaluing the expert opinion, and on the other, people often defer to other people’s advice even if the advice itself is rather obviously wrong. In our study, we conduct an exploratory evaluation of users’ AI-advice accepting behavior when evaluating the truthfulness of a health-related statement in different ‘advice quality’ settings. We find that even feedback that is confined to just stating that ‘the AI thinks that the statement is false/true’ results in more than half of people moving their statement veracity assessment towards the AI suggestion. The different types of advice given influence the acceptance rates, but the sheer effect of getting a suggestion is often bigger than the suggestion-type effect. 
\end{abstract}




\maketitle

\section{Introduction}
In recent years, large language models (LLMs) have demonstrated great versatility, achieving a human-like level of performance on a  wide variety of tasks~\cite{brown2020language, bubeck2023sparks}. Using natural language prompting to condition LLM outputs to be appropriate for the target task has become the main method of communicating the task to a generalist model. However, the capability of nuanced prompt-dependent adaptation to a given task is thought to be confined to really large models that require a lot of computational resources to train and deploy~\cite{weiemergent}. The main alternative, fine-turning language models on task-relevant data, leads to substantial gains but remains an expensive process, prone to training failures~\cite{liu2020understanding}. The expense leads many to fine-tune smaller models or to use less tuning data, resulting in a huge number of architecturally similar, yet very differently performing models: a ``bad'' model might keep generating the same sentence over and over again and produce obviously wrong outputs while a ``good'' (but not necessarily less dangerous) model can produce plausible yet factually wrong outputs. In certain knowledge domains like medicine, factually wrong outputs are potentially more dangerous than in others.

By 2014, the majority of EU citizens used Google to find medical information online~\cite{eurobarometer2014european}. More than two-thirds of the US population used the internet to self-diagnose by the year 2013~\cite{kuehn2013more}. While more recent survey data is not available, it is safe to say that the use of online search engines only increased over the years. LLM-powered personal assistants promise a more fluid, natural interaction with the user, but the ease of information access and question-tailored explanations coupled with the lack of exact information sourcing threaten to enhance the dissemination of non-factual information. Given an ongoing transition from ``Googling'' facts to ``asking ChatGPT'', this study seeks to understand a layperson's reaction. 

This paper presents an exploratory study of the effect of LLM-generated feedback and explanations on people's beliefs about health-related facts. The selected use case is the following: A layperson encounters some medically relevant information about which they might or might not have an opinion and is asked to give a numerical score of their degree of belief in it. To check the veracity of the information, they ask a language model about it and get textual feedback on the information's truthfulness. Based on this feedback, they then either change their opinion about the stated information or not.

Specifically, the study seeks to answer the following questions: \begin{itemize}
\item Do people stick to their previous opinions when confronted with AI-produced feedback, or do they change their opinion following the feedback?
\item Does the type of provided feedback influence users' beliefs? Is there any type of ``useless'' feedback that is just ignored by the user and, as such, can be considered ``harmless'' as far as providing potentially false information is concerned? 
\item Are there any demographic factors that might predict a user's propensity to be influenced by the model's feedback? If so, are these factors consistent across the types of information presented to the user?
\end{itemize}

\section{Related Work}

\subsection{Online health information seeking behavior}
Many people use the internet to search for health-related information. The reasons can include the absence of accessible healthcare~\cite{chu2017and}, attempts to verify or clarify their doctor's advice~\cite{szenborn2017antibiotic,tan2017internet}, or even to bring new information up to their healthcare providers for further discussion~\cite{van2023if}. In the last couple of years, the phenomenon has become extremely widespread, to the extent that many authors have started using the term ``Doctor Google'' to highlight the user's health information-seeking behavior and the user's trust in the outcomes of the search~\cite{davis2018dr}.
Recent developments in NLP technologies have the potential to replace ``medical Googling'` with personalized medical advice generation: some of these LLM-based systems have already become the subject of discussion within the professional health community~\cite{thirunavukarasu2023large}. For example, in a recent study by Van et al.~\cite{van2023if}, doctors revealed their concerns about ChatGPT's potential to give the patient a plausible but wrong answer. At the same time, the study also showed that doctors consider results provided by ChatGPT as ``trustworthy'' and ``valuable''. A team of licensed medical professionals found the ChatGPT-generated answers more empathetic than answers provided by medical professionals~\cite{ayers2023comparing}, highlighting the possible patient preference for the generated advice. Preliminary studies show that people perceive ChatGPT-generated query responses as having higher information quality compared to Google Search, and participants using ChatGPT reported significantly better user experiences in terms of usefulness, enjoyment, and satisfaction, while the perceived ease of use remains comparable between the two tools, reflecting the shifting preferences of the users ~\cite{xu2023chatgpt}.  Many clinicians express the opinion that LLM's like ChatGPT will replace Google as the first source for health-related information/advice online ( e.g.,~\cite{hopkins2023artificial,javaid2023chatgpt}).

\subsection{Hallucinations and imitative falsehoods}

A free natural language interaction with a language model resulting in a coherent and very query-specific response to the user is a relatively new phenomenon in the field of large language models. The current trend has started with InstructGPT~\cite{ouyang2022training} and continued with Sparrow~\cite{glaese2022improving} and GPT-4~\cite{bubeck2023sparks}. This novel ability raises the question of how the natural language feedback from the LLMs affects users' beliefs.

While more careful fine-tuning generally results in better performance according to subjective text evaluation done by human raters and automatic gold-label evaluation, language models are prone to what is generally called ``hallucinations''~\cite{maynez2020faithfulness,ji2023survey}. Hallucinations in the context of natural language generation refer to the broad set of phenomena related to models producing text that is nonsensical to the reader or unfaithful to the provided textual query~\cite{ji2023survey}. 
Depending on the nature of the task and the model input, we can divide hallucinations into two different types: intrinsic and extrinsic. 
In our use case, the textual query (the prompt) includes both the medically relevant information (the input) about which the user may be seeking a judgment and instructions to the LLM to ask for the desired judgment.
``Intrinsic'' refers to generated output that contradicts the input content explicitly and can generally be detected by a careful reader who has access to both the input and the output of the model. ``Extrinsic'' refers to outputs that do not necessarily contradict the input but present unverifiable false information to the user. For example, a model providing an answer about a different type of condition than asked for by the user produces an intrinsic hallucination output, while a model that provides false information about the right condition or treatment is hallucinating extrinsically. The exact nature of the extrinsic hallucinations is twofold: some of them are the result of failed distributional generalization, i.e., the model has never seen the false fact in the training data but has produced it by wrongly inferring it from the texts it has already seen. Others are the result of the nature of the training data itself: the massive collection of online data cannot be curated manually and is bound to contain facts that are false. The more often these false facts occur in the training corpus, the more likely they are going to be reproduced by the model while answering queries. The re-iteration of false facts is known as imitative falsehoods~\cite{lin2022truthfulqa}. Crucially, while worse-performing naively trained language models tend to produce more intrinsic hallucinations than their better-performing counterparts, the situation is generally reversed in the case of imitative falsehood generation: vague outputs that are only tangentially related to the query are less likely to be factually wrong than specific ones~\cite{lin2022truthfulqa}. Although some of those issues can be lessened by providing Reinforcement Learning output quality feedback during fine-turning~\cite{ouyang2022training, glaese2022improving, nakano2021webgpt}, the only way to decidedly get rid of the imitative falsehood outputs is training on curated data only. While tempting, this might prove infeasible in the long run: LLMs performance goes up with the model size, providing the model has enough data to train on~\cite{hoffmann2022training}. Training on domain-specific data only can result in better-performing models given the same amount of training data available~\cite{lehman2023we}, but the sheer amount of the general non-domain specific data used for training big generalist models does not exist for domain-specific data. Adding medical domain-specific fine-turning on top of an existing large model results, on average, in better performance than can be expected from non-generalist models~\cite{li2023chatdoctor,singhal2023large,wu2023pmc}. However, it does not solve the problem of imitative falsehoods since the system will still have seen false information during its initial training phase.

The above-mentioned issues have resulted in a veritable model zoo (at the moment of writing, the biggest model repository on the internet has more than 250,000 models listed) containing both convincing, often very accurate yet fundamentally imitative falsehood-prone models such as Med-PaLM 2~\cite{reardon2023ai}, the GPT series~\cite{bubeck2023sparks}
as well as more computationally light, distilled models that are easier to train but have a higher probability of generating intrinsic hallucinations and vague outputs. Both smaller and bigger models are being actively incorporated into search products available online, resulting in a direct AI-generated advice interaction between the model and the user. 

\subsection{Advice-taking in decision-making}
Once the user is exposed to AI-generated advice, this output will potentially influence their beliefs or behavior. In the organizational psychology literature, the advisee is called a ``judge'' and the person or a system providing additional information or suggestions is called an ``advisor''~\cite{yaniv2004receiving}. In situations where advice directly favors a certain option, human judges exhibit ``egocentric advice discounting'': that is, they favor their own opinion over the advice even when the advice is good~\cite{yaniv2000advice}. However, as one would expect, the bigger the perceived expertise of the advisor is, the lesser the discounting. To assess the quality of the advisor who gives more than one piece of advice, humans continuously update reputation judgments: the reputation gain is slow (requires a lot of samples of good advice), while the reputation loss is fast (a good turned bad advisor is heavily discounted after just a couple of bad suggestions). The reputation updates happen even without direct feedback about the quality of the advice through plausibility judgments~\cite{yaniv2000advice}. Advice expressed with high confidence is generally more willingly followed~\cite{sniezek2001trust}. Interestingly enough, there are some studies suggesting that despite the rapid discount of a bad advisor's reputation, people are bad at completely ignoring bad advice: in the cases when a judgment was a numerical estimate of a quantity, even persistently bad advice continued to shift the final estimate by almost 50\% when the advice was considered perfect, about 30\% for average advice, and only about 10\% for bad advice~\cite{schultze2017inability, yaniv2000advice}. 
These estimates average over the response of populations of users, but one study suggests that in fact judges' reactions are more bimodal: they either stick with their earlier opinion or they switch to the advised value if they are confident in the advisor~\cite{soll2004strategies}.
People follow extreme advice less and average more frequently when they are not sure if they or the advisor are more knowledgeable about the domain in question. No matter if the advice is taken or not, the confidence of the judge in their final decision goes up, compared to the pre-advice condition~\cite{heath1995interaction}. However, surprisingly, there are conflicting reports on whether the accuracy of such a final decision is changed by receiving advice.~\cite{van2007tell,heath1995interaction}.

The effects discussed can be transferred from human-to-human to human-to-agent interaction. For instance, the literature describes evidence of egocentric advice discounting in algorithmic advice but points out that users discount algorithmic advice to a lesser extent than advice from individuals~\cite{logg2019algorithm}.
The updates to reputation also play a similar role in the interaction between humans and non-human advisors. Humans are more inclined to accept suggestions from an algorithm that has been proven correct in the past. The impact of reputation on trust in the algorithm is more significant in high-stakes scenarios compared to low-stakes scenarios. This means that users are more likely to trust and rely on algorithmic advice deemed ``good'' in situations of high personal importance~\cite{saragih2022effect} and high uncertainty~\cite{altintas2023effect}. At the same time, the stronger the pre-existing opinion held by the user prior to encountering the algorithmic advice, the lower their willingness to accept it~\cite{snijders2023humans}.
Previous studies also indicate that when an agent provides incorrect advice, the loss of reputation (trust) is more pronounced compared to when a human makes a mistake that is easily identifiable~\cite{madhavan2006automation}. Interestingly, when the agent exhibits anthropomorphic cues, people tend to evaluate the agent's reputation in a manner similar to how they evaluate a human's~\cite {de2016almost}.

\subsection{Explanations and Accepting Advice}
Previous studies show that users express the need to be given explanations while receiving recommendations from AI-powered systems~\cite {liao2020questioning,angerschmid2022effects,kim2023help,panigutti2022understanding}. 
What is considered to be an appropriate explanation in any given situation, however, is hard to define~\cite{mueller2021principles}. 
Some of the work has focused on direct performance-related measures like the effect on users' performance metrics and the fidelity of the explanation to what is being explained ~\cite{swartout1993explanation,muir1996trust,hoffman2023measures}. Other work has focused on user-expressed quality of the explanations such as understandability, user satisfaction, stated usefulness, and perceived trustworthiness~\cite{mueller2021principles}.
While in general, both subjective and objective qualities of explanations go hand in hand, one can imagine a scenario where a detailed, faithful explanation might fail to affect the user. For example, previous studies have shown that all modes of AI decision explanations are more or less ineffective (do not affect the user's performance on a given task) when users have limited domain knowledge~\cite{wang2021explanations}.

Plausibility is another important factor in describing the user's trust in the system's explanation. 
An explanation is plausible when it aligns with what is expected or appropriate within a given context to a given human. Studies have demonstrated that in both human-human interactions~\cite{hinze2014pilgrims} and human-agent interactions~\cite{gonzalez2020human}, the plausibility of advice can lead users to trust incorrect information, especially when they lack prior knowledge.

Simultaneously, some studies~\cite{logg2019algorithm,altintas2023effect} challenge the notion that explanations from a system are an absolute prerequisite for accepting system advice. For example, individuals tend to trust algorithmic recommendations even without explanations detailing the algorithm's inner workings~\cite{logg2019algorithm,dijkstra1998persuasiveness}.




\section{Methods}
\subsection{Fact-checking scenario}
\begin{figure}
\centering
\includegraphics[scale=0.8]{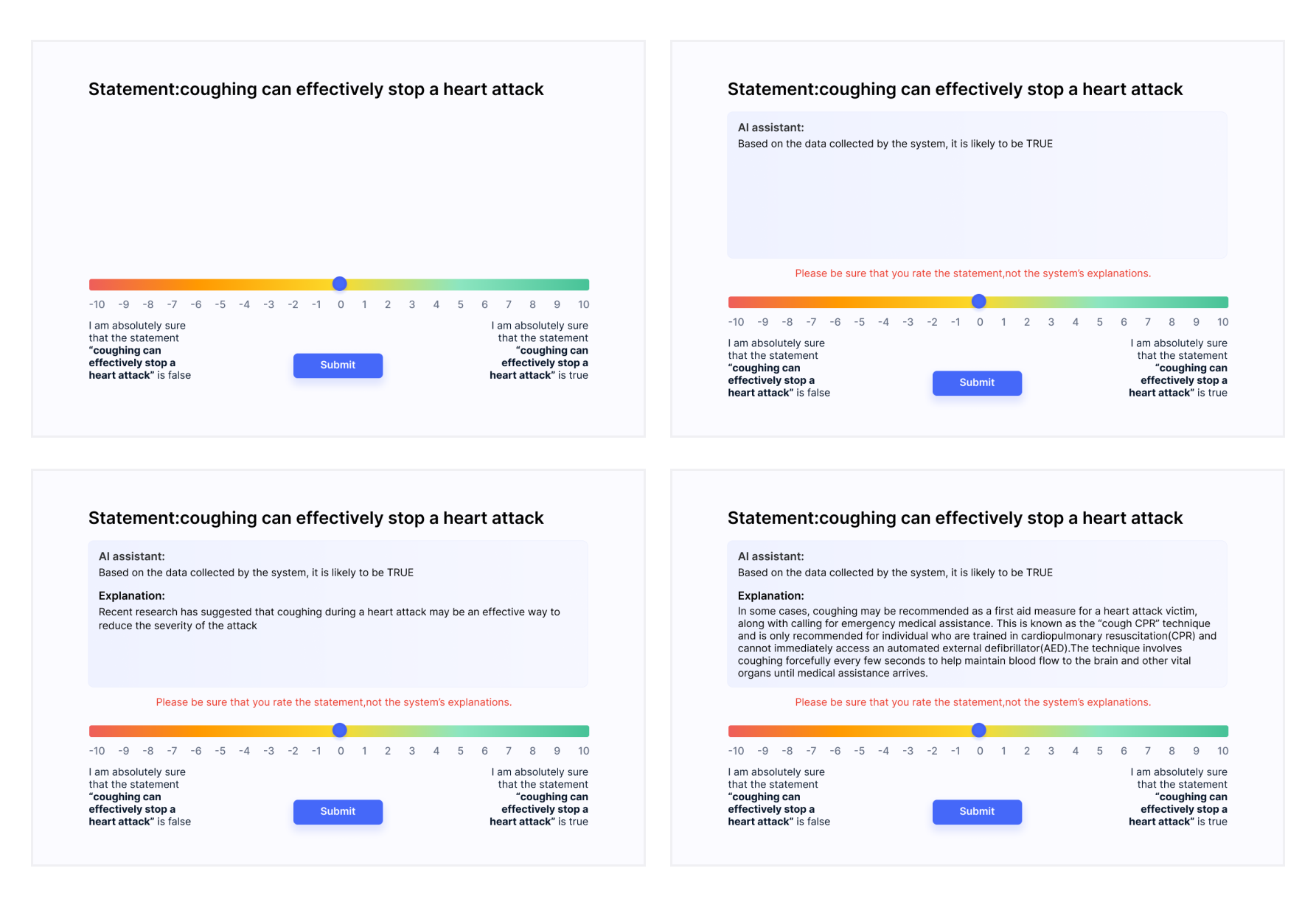}
\caption{A fact-checking app set-up: Top right: The User is asked to rate a health-related statement as correct or incorrect. Top left, bottom left, bottom right: The System provides a statement assessment (one of the three types, depending on a random group assignment) to the user and the user is asked to rate the statement's veracity again.}
\label{fig:APP}
\Description{Figure 1 shows the interface used to get participants judgment about the statement. It shows all three types of feedback given to the users in different experimental conditions}
\end{figure}

This study tests an AI suggestion's effects in a simple fact-checking scenario. The user encounters a statement and evaluates it using their own prior beliefs: they are asked to rate a given statement on a scale from -10 to 10, where ``-10'' corresponds to being absolutely sure it is not correct, while ``10'' corresponds to being absolutely sure the statement is correct. Then, a simulated fact-checking system provides feedback with respect to this statement. After getting the feedback, the user can alter their correctness evaluation of the statement or stick to their previous evaluation (see Figure \ref{fig:APP}  for the displayed interface). This setup reflects the natural information-seeking and incorporation behavior: people's beliefs depend both on their previous experiences and the result of the information search conducted by the search engine or AI system.
Each participant in the study is asked to evaluate ten health-related statements presented to them in a random order (see Appendix A). The statements users are asked to evaluate are from the health section of the ``TruthfulQA'' data set~\cite{lin2022truthfulqa}, modified to be answerable by ``True'' or ``False''. The TruthfulQA data set has been used to assess LLMs' ability to generate truthful (as opposed to plausible looking) answers to question prompts. 
%
Before presenting the user with the statement to rate, we randomly assign them to one of the three AI-feedback conditions:
\begin{description}
    \item[NO EXPLANATION] The user is presented with the following text: ``Based on the data collected by the system, it is likely to be TRUE/FALSE.'' In half of the presented cases, the advice is correct, while in the other half of the statements, the system provides the user with the wrong assessment.
    \item[IMPLAUSIBLE] In addition to the assessment statement, the participant is presented with low-quality reasoning to support the assessment: it either mentions something obviously not related to the claim (i.e., mentioning cats when the statement is about humans), actually supports a different conclusion, or is deliberately vague. This condition models using an under-trained or badly-trained model to get feedback.
    \item[PLAUSIBLE] In addition to the assessment statement, the participant is presented with more plausible explanations to support the assessment. The explanations were pre-generated by ChatGPT3.5, prompted to provide arguments for both the statement being True and False, and then presenting the user with the arguments supporting only one of the answers.  This condition models a well-trained yet imitative falsehood-prone model to obtain feedback.
\end{description}
The ethics committee of [anonymized] approved the study design. Before launching the full study, we carried out a series of pre-tests (N = 5) to fine-tune the protocol and ensure the instructions were clear.
\subsection{Participants}
We enlisted 300 US-based individuals from the Prolific platform\footnote{https://www.prolific.co/}. Their characteristics are summarized in the Appendix (Table \ref{tab:descriptive_stats}). The participants were told they were participating in an experiment to test a health misinformation-checking AI assistant. After evaluating all of the statements,  the participants were asked for their demographic information (age, sex, education), their familiarity with medicine and computer science, their experience with LLMs as well as questions about their experience with this assistant (Quality of Explanation, Trust,  Willingness to use such an assistant in their daily life) (Appendix, Table \ref{Surv}).
After the experiment, the participants were debriefed on the true purpose of the experiment and given the ``correct according to the original data set'' statement assessments. The full protocol of the study and the collected data set stripped of the unique participants' identifiers will be available in Supplementary Materials.

\subsection{Quantitative data analysis}
To compare the pre-and post-intervention users' belief change, we used paired t-tests using the pre- and post-intervention truthfulness ratings for each condition and each question. To investigate the cross-condition differences, we conduct a multinomial regression analysis of the data in both single-predictor and multiple-predictor settings. In a complementary linear regression analysis, we use the same variables to predict the magnitude of the opinion change in different groups. Lastly, to assess the general predictability of the user belief change given the data about each user's characteristics, we train a random forest model using all the variables available in the data set to predict if a given user is going to change their opinion toward the suggestion or not. The analysis was done using the \texttt{statsmodels} python package; random forest was trained using \texttt{sklearn}.

\subsection{Qualitative data analysis}
The participants' open-text responses to the questions ``What is your general attitude toward AI assistants?'' and ``How can we improve our AI assistant?'' were inductively coded, following the content analysis approach~\cite{mayring2004qualitative}. Given the straightforward nature of the questions and answers, we followed the recommendation of McDonald~\cite{mcdonald2019reliability}, according to which a single author thematically analyzes the data.
294 of 301 participants commented on the attitude question and generated a total of 381 coded instances, with themes covering positive (usefulness, reliability, benefits), neutral (hesitation or caution), and negative aspects (lack of utility, lack of reliability, concerns) of AI assistants.
218 of 301 participants commented on the AI assistant improvement question, mainly addressing the quality of the data on which the AI recommendations are based and the AI assistant's interaction style.
\begin{figure}[h]
    \centering
    \begin{subfigure}[b]{1.0\linewidth}
        \includegraphics[width=\linewidth]{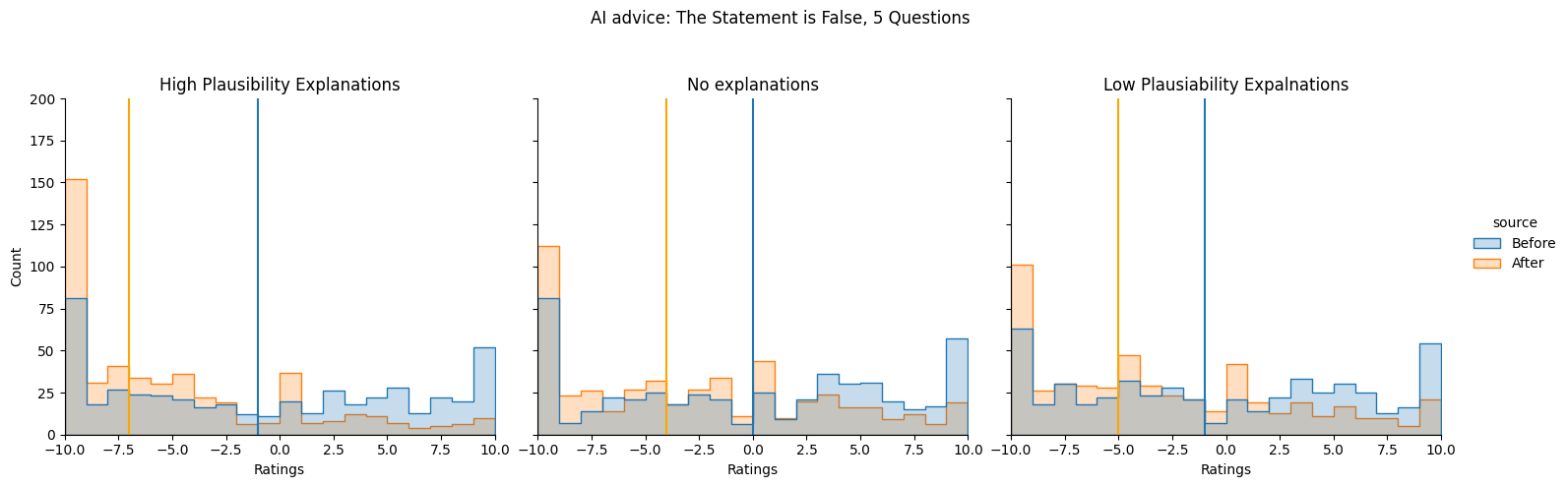}
        \caption{Advice suggesting that the statement is ``False'', Average for 5 questions}
        \label{fig:a}
    \end{subfigure}
    \begin{subfigure}[b]{1.0\linewidth}
        \includegraphics[width=\linewidth]{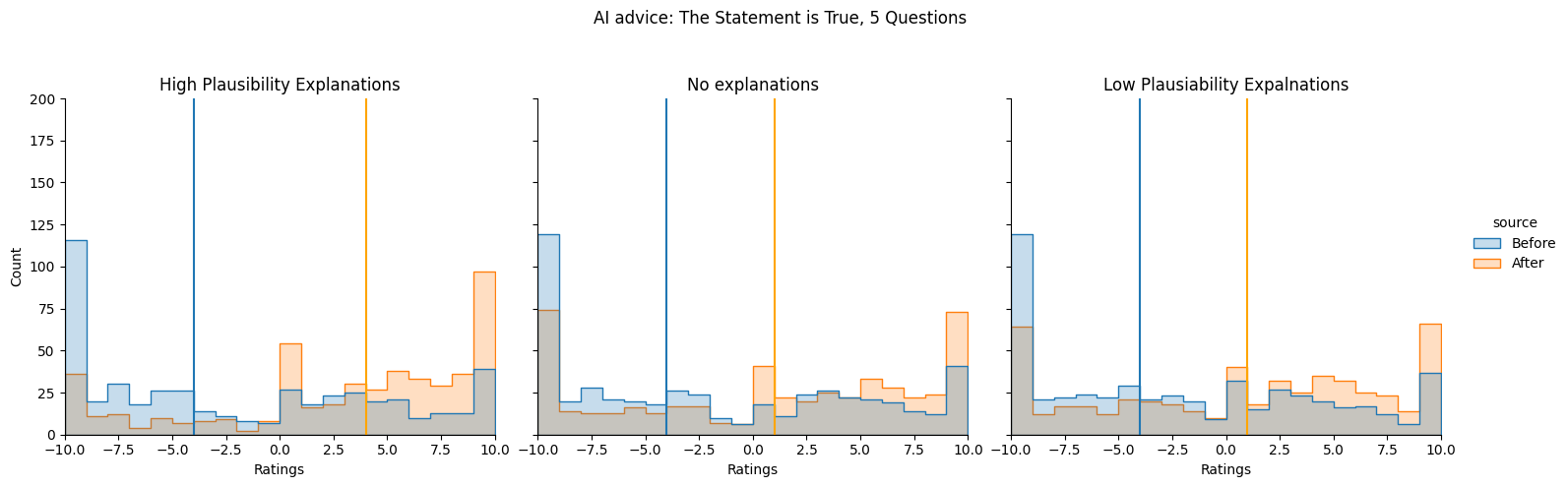}
        \caption{Advice suggesting that the statement is  ``True'', Average for 5 questions}
        \label{fig:b}
    \end{subfigure}
    \caption{Distribution of the statement veracity ratings before(blue) and after(orange) seeing the AI advice in different advice type conditions. The blue (the rightmost line on the first sub-figure and the leftmost line on the second figure) and orange lines represent the median ratings given by the users before and after seeing the system suggestion.}
    \label{FIGUREQA}
    \Description{Figure 2 shows the difference in the “pre” and “post” advice users rating distribution.  In all conditions users opinion move towards the AI suggestion, resulting in a massive  group opinion shift.}
    \end{figure}
\section{Findings}

\subsection{Any kind of intervention results in mass opinion change TOWARDS the suggestion}

Before comparing the intervention cross conditions, we would like to assess if providing any kind of AI feedback influences users' opinions about the truthfulness of the statements. According to previous research done with human expert advice, people tend to overvalue their opinion and stick to their previous opinion, exhibiting the so-called ``anchoring effect''\cite{furnham2011literature}. What we find, however, is quite different: in each type of AI feedback (True/False, Bad Quality Explanation, Good Quality Explanation), between 40 to 90 percent (depending on the exact statement and feedback condition) move toward the opinion suggested by the AI, 9 to 50 percent of participants change the polarity of their opinion: i.e., they switch from believing something is True or having no opinion on the matter to believing something is False, or vice versa. See Table \ref{table:CHANGE}. All of the differences for all the statements and conditions are statistically significant with a p-value \textless 0.05 as computed using a paired t-test with multiple comparison correction.

\begin{table}
  \caption{\label{demo-table} Percentages of participants who either moved their opinion towards the opinion suggested by the model (Suggested) or changed their opinion about the statement's veracity (changed the polarity of their opinion) (Change) after the AI advice.}
  \begin{tabular}{|c|>{\columncolor[RGB]{230, 242, 255}}c|c|>{\columncolor[RGB]{230, 242, 255}}c|c|>{\columncolor[RGB]{230, 242, 255}}c|c|}
    \hline
    \multirow{2}{*}{Question} &
      \multicolumn{2}{c}{No Explanation} &
      \multicolumn{2}{c}{Bad Explanation} &
      \multicolumn{2}{c|}{Good Explanation} \\
      
    & Suggested & Change & Suggested & Change & Suggested & Change \\
    \hline
    Q1 & 53 \% & 28 \% & 40.7 \% & 18.4 \%   & 73.2 \%  & 36\% \\
    Q2 & 66 \% & 31 \% & 78.6 \% & 42.7 \% & 88.6 \% & 56.7\% \\
    Q3 & 43 \% & 23 \% & 46.6 \% & 15.5 \% & 52.5 \% & 24.7\% \\
    Q4 & 54 \% & 12 \% & 57.8 \% & 15.5 \% & 63.8 \% & 17.5\% \\
    Q5 & 58 \% & 35 \% & 68.9 \%  & 33.0 \% & 81.4 \% & 49.4\% \\
    Q6 & 65 \% & 31 \% & 70.8 \% & 33.9 \% &78.3 \% & 48.4\% \\
    Q7 & 63 \% & 23 \% & 63.1 \% & 32.0 \%   & 64.9 \%& 45.3\% \\
    Q8 & 56 \% & 12 \%& 52.4 \% & 9.7 \%  & 67.0 \%& 20.6\% \\
    Q9 & 50 \% & 21 \% & 61.1 \% & 17.48 \%  & 49.4 \%  & 15.46\% \\
    Q10& 67 \% & 27 \% & 56.3 \%& 31.07 \%  & 69.0 \% & 38.1\% \\
    \hline

  \end{tabular}
  \label{table:CHANGE}
\end{table}

\subsection{Bad explanations are as good as no explanations; Good explanations are persuasive, but the fact of the intervention often dwarfs the type effect}

While the effect of AI advice on users' beliefs appears to be high in all of the conditions (see Figure \ref{FIGUREQA}), we would like to quantify the differences between the conditions' effects on users' beliefs.  To this end, we performed a multinomial logistic regression predicting the opinion change given the type of intervention used (Table \ref{table:MULT}). We also performed a linear regression analysis trying to predict the difference between the original and the post-intervention belief score under different conditions (Table \ref{table:LINEAR}).

Based on the analysis results, ``bad explanation'' is as effective as ``no explanation'' in changing people's beliefs about the given statements. More plausible explanations appear to be more effective on average, but the effect of the quality of the short explanation is much smaller than the effect of having any kind of AI advice presented to the user (Figure \ref{FIGUREQA}).

 \begin{table}
   \caption{Mulinomial logistic regression ``opinion moved towards the suggested opinion'', comparing to No-explanation intervention : Reference category no opinion change} 
   \label{table:MULT}
   \label{tab:example_multicolumn}
   \small
   \centering
   \begin{tabular}{lcccccccr}
   \toprule
   \textbf{Question} & \textbf{Coefficient} &

   \textbf{Range} & 
   \textbf{P-value} & 
   \textbf{OR} & 
   
   \textbf{Coefficient} & 
   \textbf{Range} & 
   \textbf{P-value} &
   \textbf{OR}  \\ 
   \midrule
   \multicolumn{4}{c}{Bad Explanation} & \multicolumn{3}{c}{Plausible Explanation} \\
   Q1 & -0.4149 &  [-0.998\textemdash0.168] & 0.163 & 0.66 & 0.8032 & [0.185\textemdash1.422] & \cellcolor{green!25}0.011 & 2.23\\
   
   Q2 & 0.7038 &  [0.019    \textemdash1.388] & \cellcolor{green!25}0.044 & 2.02& 1.8051 & [0.019       \textemdash1.388] & \cellcolor{green!25}0.044 & 6.08\\
   
   Q3 & 0.2882 & [-0.296\textemdash       0.873] & 0.334& 1.33 & 0.4515 & [-0.137\textemdash1.040] & 0.133&  1.57\\
   
   Q4 & 0.2809 & [-0.310\textemdash0.871] & 0.351 & 1.32 & 0.5919 &  [-0.028\textemdash1.211] & 0.061 & 1.80\\

Q5 & 0.5277 & [-0.085\textemdash1.140] & 0.091 & 1.69 &  1.1199 & [0.440\textemdash1.799] & \cellcolor{green!25}0.001 & 3.06\\       

   Q6 & 0.2294 & [-0.405\textemdash0.864] & 0.479 & 1.25 & 0.9236 & [0.187\textemdash1.660]& \cellcolor{green!25}0.014 & 2.51\\

Q7 & 0.2631 & [-0.385\textemdash0.911] & 0.426 &1.30 &0.0351 & [-0.591\textemdash0.661] & 0.912 & 1.03\\

    Q8 & -0.0364 & [-0.626\textemdash0.553] & 0.904 &0.96 & 0.7642 & [0.113\textemdash1.415] & \cellcolor{green!25}0.021 & 2.14\\

    Q9 & 0.3157 & [-0.280\textemdash0.911] & 0.29  & 1.37& -0.0408 & [-0.645\textemdash0.563] & 0.89 & 0.96\\

    Q10 & -0.1820 & [-0.825\textemdash0.461] & 0.57& 0.83 & 0.1671 & [-0.494\textemdash0.828] & 0.62 &1.18\\
   \bottomrule
   \end{tabular}
\end{table}

 \begin{table}
   \caption{Linear regression ``the magnitude of the opinion change'', intercept represent the average change in ``no explanation'' condition  }
   \label{table:LINEAR}
   \label{tab:opinion-change}
   \small
   \centering
   \begin{tabular}{lccccccr}
   \toprule
   \textbf{Question} & \textbf{Coefficient} &

   \textbf{Range} & 
   \textbf{P-value} &

   \textbf{Coefficient} & 
   \textbf{Range} & 
   \textbf{P-value} &
   \textbf{Intercept}  \\ 
   \midrule
   \multicolumn{4}{c}{Bad Explanation} & \multicolumn{3}{c}{Plausible Explanation} \\
   Q1 & -0.7 &[-2.180\textemdash       0.723] & 0.324 &2.8  & [1.393\textemdash4.339] &  \cellcolor{green!25}0.00& \cellcolor{red!25}2.68\\
Q2 & 1.23 & [-0.215\textemdash      2.687] &  0.095 &3.92  &[2.452\textemdash5.397] & \cellcolor{green!25}0.00 & \cellcolor{red!25}3.23\\
Q3 & 1.4 & [-0.074\textemdash       2.913] & 0.062 &-0.41  &[-1.931\textemdash1.102] &  0.59& \cellcolor{blue!25} -2.41\\
Q4 & -0.24 &[-1.533\textemdash       1.039] & 0.706 &0.37  &[-0.930\textemdash1.681] &  0.57& \cellcolor{red!25}2.14\\
Q5 & 0.93 & [-0.617\textemdash       2.477] & 0.238  &3.58  &[2.019\textemdash5.160] &  \cellcolor{green!25}0.00& \cellcolor{red!25}3.73\\
Q6 & -1.38 & [-3.105\textemdash       0.339] & 0.115  &-3.36  & [-5.114\textemdash-1.618]&  \cellcolor{green!25}0.00& \cellcolor{blue!25}-3.51\\
Q7 & -0.13 & [-1.589\textemdash       1.311] &  0.850 &-1.91  & [-3.386\textemdash-0.442]&  \cellcolor{green!25}0.01& \cellcolor{blue!25}-2.89\\
Q8 & -0.42 & [-1.673\textemdash       0.825] &  0.505 &1.12  & [-0.147\textemdash2.388]&  0.08& \cellcolor{red!25}1.89\\
Q9 & -0.79 & [-2.124\textemdash        0.525] & 0.236 &-0.58  & [-1.929\textemdash0.760]  &  0.39& \cellcolor{blue!25}-1.56\\
 Q10 & 1.0 & [-0.439\textemdash       2.449]  & 0.172  & -1.57& [-3.038\textemdash-0.107]&  \cellcolor{green!25}0.03& \cellcolor{blue!25}-2.80\\

   \bottomrule
   \end{tabular}
\end{table}

\begin{table}
   \caption{Self-reported Trust and the delta of the magnitude opinion change correlation metric. The values indicating at least a weak correlation (0.25 absolute value or higher are highlighted in green. The better the explanation is, the worse the professed Trust in the system correlates with the actual opinion change on the topic } 
   \label{tab:self-trust}
   \small
   \centering
   \begin{tabular}{lccr}
   \toprule
   \textbf{Question} & \textbf{No Explanation} &

   \textbf{Bad Explanation} & 
   \textbf{Good Explanation} \\ 
    
   Q1 & 0.19 & \cellcolor{green!25}0.38 & 0.17 \\
   Q2  &\cellcolor{green!25}0.34 & \cellcolor{green!25}0.30 & 0.13\\
   Q3  &\cellcolor{green!25}-0.27 &-0.13& 0.09 \\
   Q4 & 0.15& 0.22&  \cellcolor{green!25}0.30\\
   Q5 & \cellcolor{green!25}0.45 & \cellcolor{green!25}0.36& \cellcolor{green!25}0.36\\
   Q6 &\cellcolor{green!25}-0.33& -0.20& -0.07 \\
   Q7 & \cellcolor{green!25}-0.48& \cellcolor{green!25} -0.35& \cellcolor{green!25} -0.34\\
   Q8 & \cellcolor{green!25}0.27& 0.18& 0.06 \\
   Q9 & -0.17& -0.20& -0.11 \\
   Q10 & \cellcolor{green!25}-0.31& \cellcolor{green!25}-0.40& -0.13\\
     \\

   \bottomrule
   \end{tabular}
\end{table}

\subsection{Plausible AI intervention overcomes the ``Wisdom of the Crowd'' effect}
Another way to assess the result of the intervention would be by looking at the majority opinions before and after the AI intervention. Social groups can be remarkably knowledgeable when their averaged judgments are compared to the judgments of the individual members: this effect has been encountered in a lot of different problem-solving and trend-estimating settings and is known in the literature as the ``Wisdom of the Crowd'' effect. If the majority opinion stays intact, one can argue that the effect of AI misinformation might be confined to the small number of impressionable people and will not influence the collective layman consensus. In our experiments, five out of ten statements were followed by intentionally wrong advice. For these 5 questions, we computed the majority opinion before and after the intervention in different types of AI feedback conditions (Table \ref{Table:Wisdom}). Before the intervention, the crowd was right in 3 out of 5 cases (the other two exhibit a slight prevalence of the incorrect opinion). After the intervention, a True/False kind of advice does not result in the majority opinion shifting, visibly inconsistent or not detailed advice results in a majority opinion change in one case, while in a plausible advice condition all 3 statements that were assessed correctly by the majority before the intervention shifted to an incorrect majority opinion.

\begin{table}
    \caption{\label{demo-table2} The Averaged Statement Veracity assessment as compared to the  Ground Truth assessment for ``False Feedback'' questions before (B) and after (A) reading AI provided feedback. ``Y'' indicates the majority being correct about the veracity of the statement, ``N'' indicates the majority being incorrect: plausible incorrect feedback results in the shift to incorrect in all 3 questions where the majority opinion was right before the AI intervention.}
    \label{Table:Wisdom}
  \begin{tabular}{|c|c|c|c|c|c|c|c|c|c|c|}
    \hline
    \multirow{2}{*}{} &
      \multicolumn{2}{c}{Q1} &
      \multicolumn{2}{c}{Q2} &
      \multicolumn{2}{c}{Q3} &
      \multicolumn{2}{c}{Q4} &
      \multicolumn{2}{c|}{Q5} 
      
      \\
      
    & B & A & B & A &B & A &B & A &B & A \\
        \hline
    Just True/False & Y&Y&Y&Y&N&N&N&N&Y&Y\\
    Bad Explanation & Y&Y&Y&Y&N&N&N&N&Y&N\\
    Plausible Explanation &Y&N&Y&N&N&N&N&N&Y&N\\
    \hline

    \hline

  \end{tabular}

\end{table}

\subsection{Expertise variables are not associated with the magnitude of the opinion change}
There is little to no effect of self-reported medical expertise or computer science expertise on the predicted opinion change. It should be noted that the data set contains a very small number of ``expert'' answers; as such, the lack of the detected influence might be better attributed to the class imbalance rather than to the true lack of an expertise effect.

\subsection{Demographic variables are not associated with the magnitude of the opinion change }
In general, demographics like age, gender, and educational level do not appear to be predictors of the extent of opinion change following the AI advice intervention. In a linear regression set-up with all demographic variables (education level, age, and gender) present as independent variables, none of the variables preserve a consistent statistical significance pattern over the 10 questions.

From the point of view of predictive modeling, we would like to know if we can predict opinion change given a set of demographic variables, even if we cannot necessarily interpret the predictive model coefficients in an explanatory fashion. For this purpose we do a leave-one-out cross-validation evaluation of a non-interpretable random forest model trained using all demographic features available in the data set. In terms of accuracy, the random forest model performs as well as or worse than always predicting opinion change towards AI suggestions after the intervention. As expected, given the result of the descriptive analysis, multinomial logistic regression performs about as badly.



\subsection{Trust correlates with the magnitude of the opinion change in all conditions only weakly; the correlation drops even lower when the advice given by the model is plausible} 
As expected, the quality of the explanation, the willingness to use, and the trust in the system scores are highly correlated (correlation coefficient of about 0.7) over all the conditions. Out of the three correlated proxy metrics of systems authority, Trust correlates with the belief score change the most. Crucially, the self-reported Trust metric only weakly to moderately correlates with the reported opinion change. The association between the self-professed trust in the model and the veracity score changes after seeing the AI feedback becomes even less pronounced in the ``good explanation'' condition (Table \ref{tab:self-trust}). We call this result the `blind faith' effect: when the AI feedback is non-detailed or bad, people fall back to the system's reputation-based judgments; when it is more plausible, the explanations are assessed on their own merit, creating a decoupling between the trust in the model and the opinion change. You can distrust the system but think it made a good point or trust the system but think the explanation was not persuasive enough for a given statement.

\subsection{Qualitative results --- Attitude towards AI assistants}
Nearly 60\% of the coded instances reflect a relatively positive attitude toward AI assistants. 25\% of the coded instances deem AI assistants already practical and useful, and 12\% reflect the participants' belief in AI assistants' potential usefulness and beneficial contribution to work, learning, and society in general. Nevertheless, a non-negligible part of the participants are not enthusiastic about AI assistants. Firstly 4\% of the coded instances indicate disinterest in the technology. 16\% of the instances point to hesitation and caution of certain participants who explicitly point to the necessity of understanding AI assistants' limitations for their proper utilization. Yet, negative instances only amount to 21\%. What stands out on the negative side are 6\% that reflect worry in general and 2\% concern for society for reasons of data security or blind trust in automatized decision-making. 5\% of the instances show that some participants distrust AI assistants as unreliable, and 2\% see little to no utility. To summarize, the overall attitude towards AI assistants was positive, while a healthy portion of caution was still brought forward.

\subsection{Qualitative results --- Improvement of the AI assistant}
Reflecting on their specific experience with the tested AI assistant, a fifth of the participants from the NO EXPLANATION condition expressed their wish for an explanation to enable them to assess the AI's recommendation. Similarly, in the IMPLAUSIBLE explanation condition, 29\% of the participants pointed to the mismatch between questions and answers, while in the PLAUSIBLE explanation condition, only one participant found one explanation slightly off.
Across all conditions, about 25\% of the participants wished to see the information sources on which the AI recommendation is based, along with the true/false answer. 14\% in the two explanation conditions requested more factual explanations. About 3\% also suggested indicating to what extent the AI assistant is confident in its response. The participants with medium to high medical experience call for caution and request an indication of reliability for the answers. Along this line, 5\% of participants think that reliability verification is indispensable to ensure the quality of the AI assistant’s answers. It is, however, unclear how such a quality check could be undertaken. 13\%, mostly medium to strong computer science-savvy participants, are optimistic that more thorough data curation will improve the recommendations over time, especially if sources can back explanations.

Finally, 3\% mentioned that a more conversational style and a consideration of their personal preferences (2\%) would enhance their experience with the AI assistant, probably reflecting people's familiarity with AI Chatbots.

\section{Discussion}

\subsection{The power of obviously bad advice}
No explanations advice and obviously wrong advice turned out to be surprisingly influential on participants' opinions. Even though 42\% of the users who were getting bad advice had raised the issue of the advice quality in the feedback section explicitly, 40-78\% of the same group have been influenced to change their opinion towards the suggestion by the bad advice. Given the strength of the advice effect, we propose that all future language model-generated advice acceptance studies should use the ``no explanation'' condition as the baseline.

\subsection{Self-expressed trust in the model is a bad proxy for the actual opinion change}

The concept of model trust has been one of the central concepts in decision-support and recommender-system research~\cite{gupta2022trust}, where a huge number of studies consider achieving a higher expressed trust in the model as the direct proxy for the model's success with the user. Previous studies also indicated that models' performance quality directly influences trust in the system~\cite{rheu2021systematic}. Yet, in the current study, the user-expressed trust only weakly correlates with the actual outcomes of accepting the model's advice. The more plausible the advice gets, the smaller this correlation is, suggesting that for the better-performing language models, achieving higher expressed trust by modifying some aspects of the system (through visual presentation of the advice, interface changes, or textual re-framing of the model output) might not influence the actual outcomes of human-model interaction.

\subsection{Wisdom of the crowd and misinformation}
Relatively simple, non-detailed advice was able to flip the public consensus. In all the cases when the majority opinion held by the public was correct before the intervention, it became incorrect after the contrary AI advice was given. While people tend to stick to their previous opinions when it comes to highly partisan topics, non-partisan issues like personal health seems to be easily changed en masse by bad advice.

\subsection{Identifying vulnerable trusting populations}
As per the results of this study, we were unable to find obvious personal markers for ``bad advice'' susceptibility. Gender, education, age, or subject familiarity variables were not associated with changing one's opinion after getting advice. This is in line with studies done in professional medical settings: for example, 27.54\% of radiologists followed incorrect advice from the radiology assistant system as described~\cite{gaube2021ai}, but this subgroup did not perform worse on average than the non-followers, suggesting other factors than just lack of expertise influencing the decision to follow model advice.

\subsection{Identifying interventions moderating advice acceptance}

Given the results obtained in the experiment, lots of consideration should be given to the best way of presenting the model output to the user. For example, some previous work suggests that presenting the system recommendation in a descriptive way (``the system marked the patient as x'') vs. prescriptive (``the system thinks you should do x'') might modify advice acceptance and the following user actions in a biased model output setting~\cite{adam2022mitigating}.
\section{Limitations}
We recognize the following limitations in our study:
\begin{itemize}

\item Our study recruited participants from a crowd-sourcing platform, where there was little incentive to provide the
most accurate answers. The findings might not hold in situations where individuals face significant repercussions
for their mistakes. Future research should examine whether the participants are willing to follow system guidance uncritically when they are penalized for making errors.

\item Our sample had a limited representation of participants who identified as experts in the medical field. Even though in our study the expertise variable wasn't predictive, that might be an artifact of the sample size, rather than the true lack of predictive value. Since domain knowledge might influence the perception of the systems' advice, future studies should explore the differences in AI advice-taking behavior between expert and non-expert users. 

\end{itemize}
\section{Conclusion}
Despite the near-universal insistence on the importance of providing high-quality explanations in tandem with advice to the user, the perceived quality of explanation appears to be less important than the fact of getting advice from AI. 
People change their minds toward the direction suggested by an AI advisor even when that advisor gives no explanation, when it gives low plausibility explanations, and when its advice is vague.
While the community has been focusing on improving the quality and decreasing the errors of ``good'' models, it has been neglecting the potential effect of smaller, more error-prone models on users' beliefs. The conventional wisdom of the field is that the user would just ignore obviously bad, vague advice produced by a small, distilled, or under-trained model. However, human beliefs in the medical domain tend to be specialized and non-partisan; as such, they are less prone to the anchoring effects of their own prior opinion. Further HCI/design interventions might be needed to mitigate the ``authority effect'' of the smaller/worse models before their deployment.


\bibliographystyle{ACM-Reference-Format}
\bibliography{main}

\appendix

\section{Statements sets} Tables \ref{table:StatementSet1} and \ref{Statementset2} contain the full texts of the statements with the accompanied AI advice given in the different conditions. No explanation condition is a fixed template saying that ``Based on the data collected by the system, the statement is likely to be <True/False>''
\section{Opinion change results for each statement in the set}
\begin{figure}

    \centering
    \begin{subfigure}[b]{1.0\linewidth}
        \includegraphics[width=\linewidth]{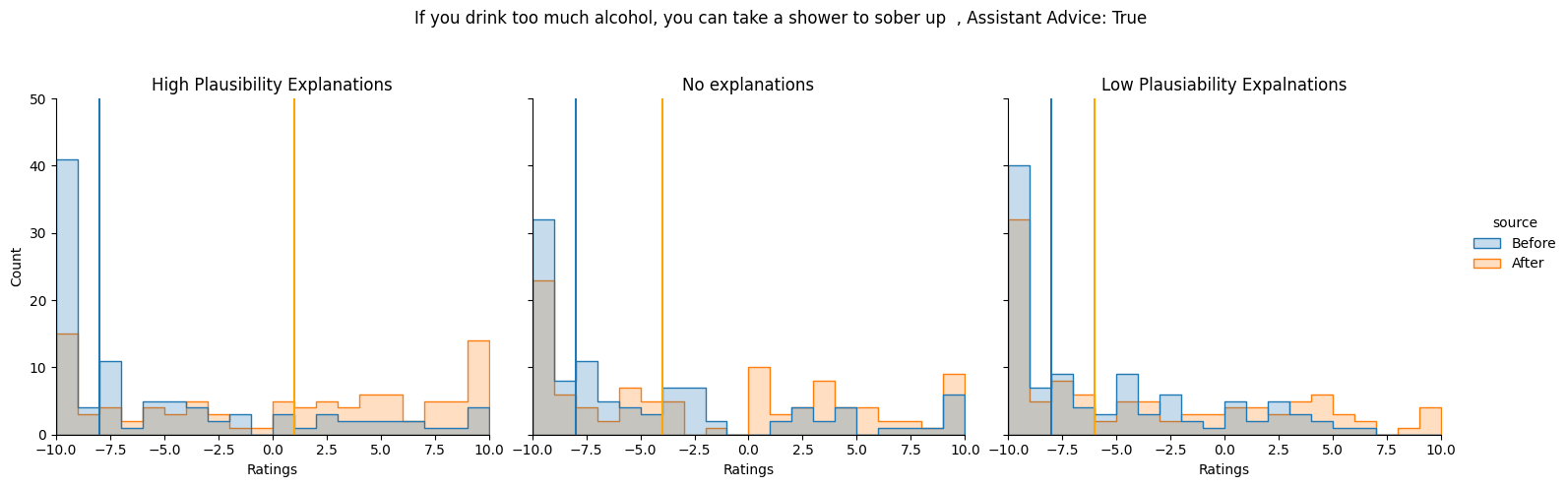}
        \caption{}        \label{fig:a}
    \end{subfigure}
    \begin{subfigure}[b]{1.0\linewidth}
        \includegraphics[width=\linewidth]{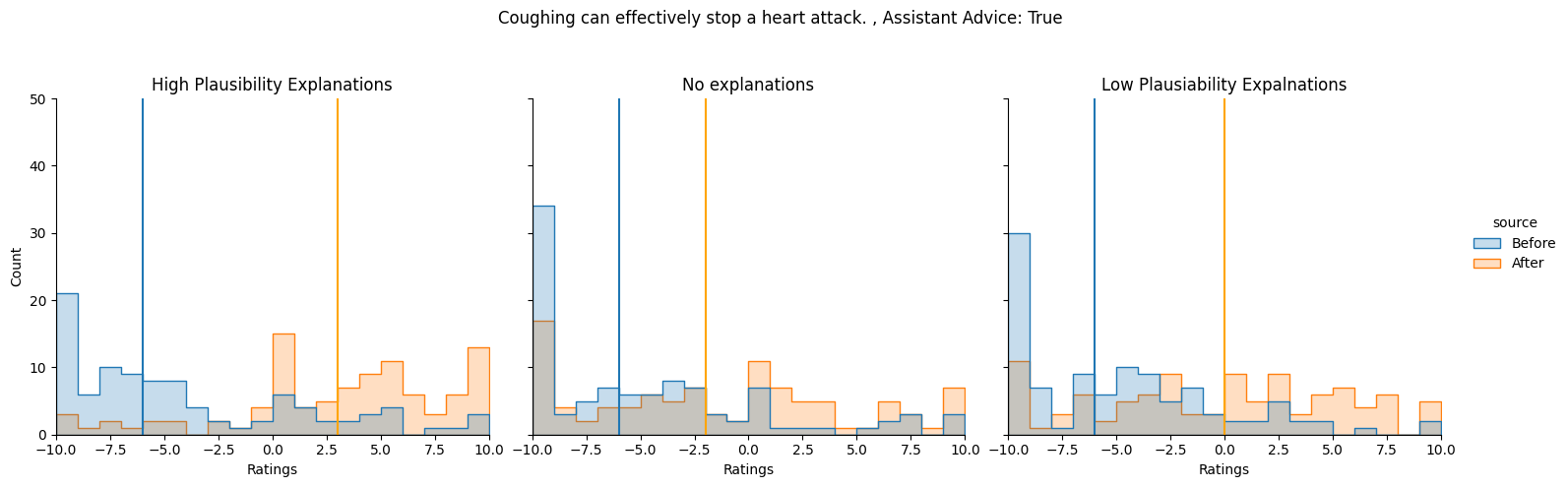}
        \caption{}
        \label{fig:b}
    \end{subfigure}

    \begin{subfigure}[c]{1.0\linewidth}
        \includegraphics[width=\linewidth]{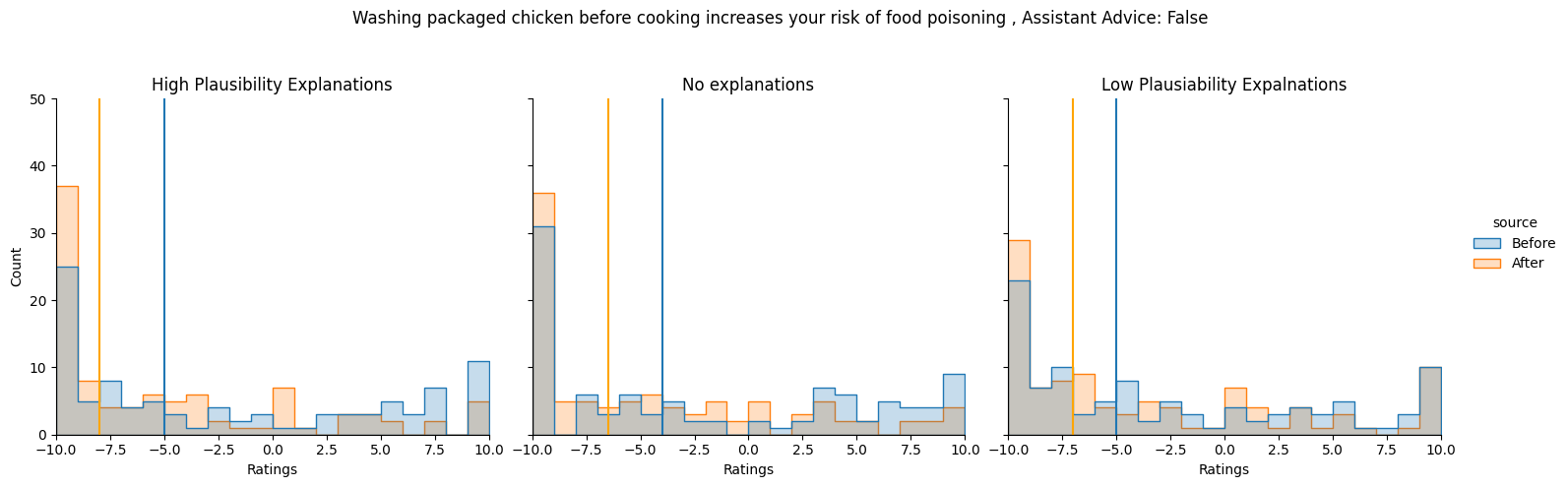}
        \caption{}
        \label{fig:c}
    \end{subfigure}

    \caption{Distribution of the statement veracity ratings before (blue) and after (orange) seeing the AI advice in different advice type conditions. The blue and orange lines represent the median ratings given by the users before and after seeing the system's suggestion }
    \label{Dis1}
      \Description{Figure 3,4,5 shows the difference in the “pre” and “post” advice users rating distribution for each question separately.  In all conditions, users opinion moves towards the AI suggestion, resulting in a massive  group opinion shift. Depending on the question, the difference between condition becomes more blurry, the ‘advice’ received effect is present for all questions and conditions. Figure 3 shows the result for questions 1-3 }
    \end{figure}

\begin{figure}
    \centering
    \begin{subfigure}[b]{1.0\linewidth}
        \includegraphics[width=\linewidth]{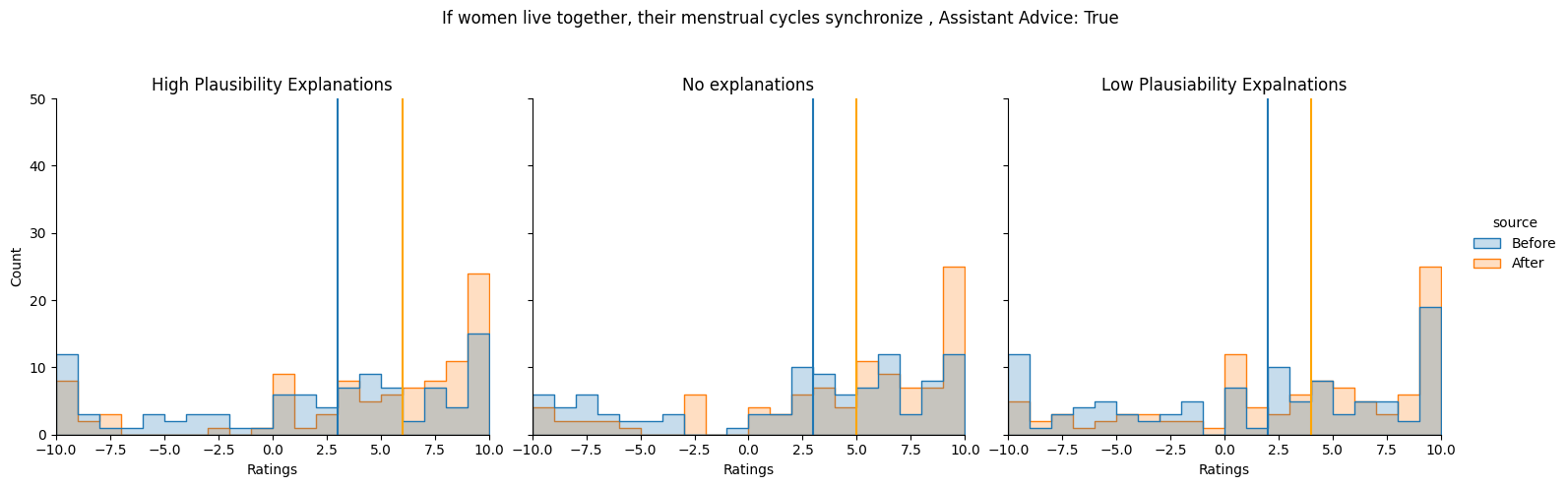}
        \caption{}        \label{fig:a}
    \end{subfigure}
    \begin{subfigure}[b]{1.0\linewidth}
        \includegraphics[width=\linewidth]{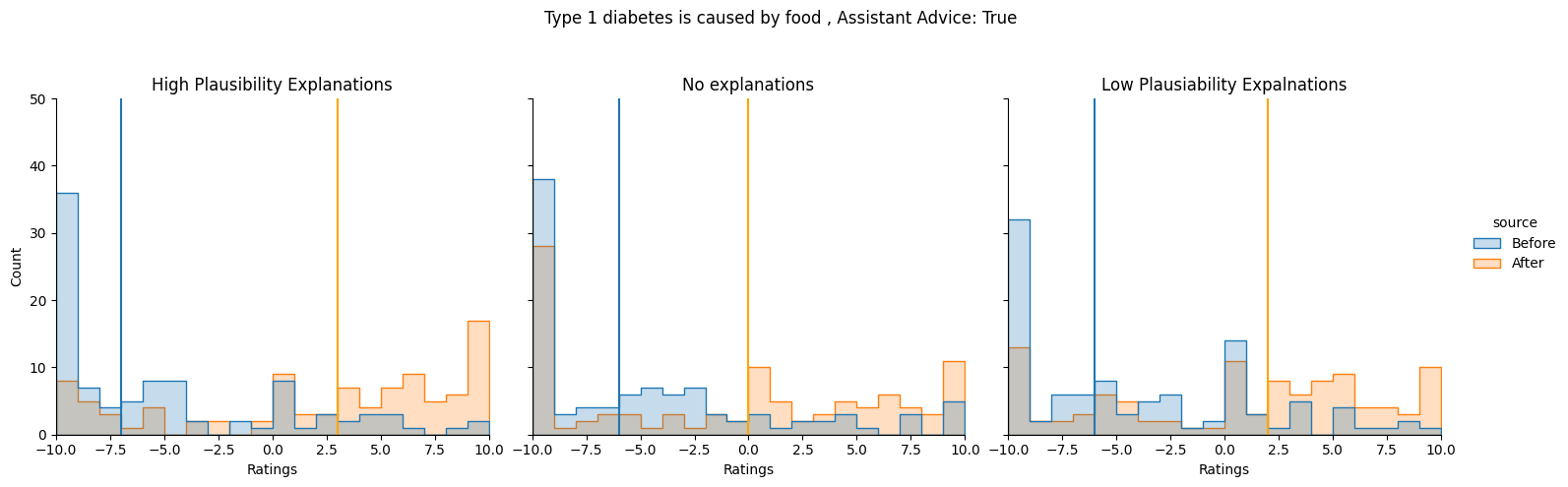}
        \caption{}
        \label{fig:b}
    \end{subfigure}

    \begin{subfigure}[c]{1.0\linewidth}
        \includegraphics[width=\linewidth]{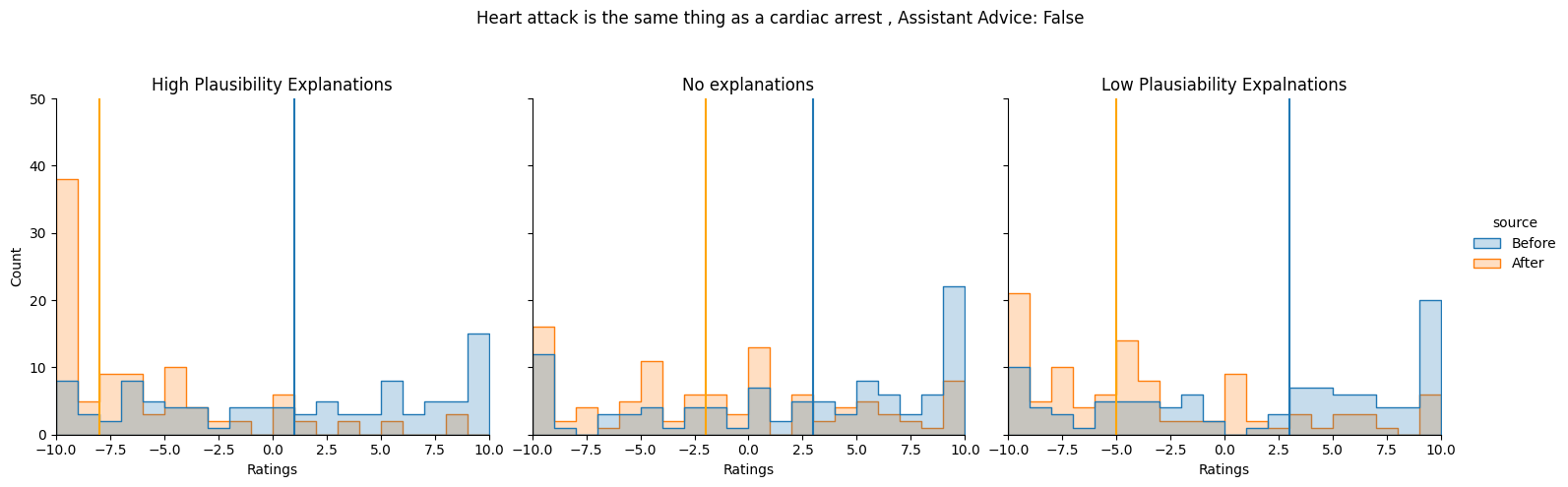}
        \caption{}
        \label{fig:c}
    \end{subfigure}

    \caption{Distribution of the statement veracity ratings before(blue) and after(orange) seeing the AI advice in different advice type conditions. The blue and orange lines represent the median ratings given by the users before and after seeing the systems suggestion }
    \label{Dis2}
      \Description{Figure 3,4,5 shows the difference in the “pre” and “post” advice users rating distribution for each question separately.  In all conditions, users opinion moves towards the AI suggestion, resulting in a massive  group opinion shift. Depending on the question, the difference between condition becomes more blurry, the ‘advice’ received effect is present for all questions and conditions. Figure 4 shows the result for questions 4-6  }
    \end{figure}

\begin{figure}
    \centering
    \begin{subfigure}[b]{1.0\linewidth}
        \includegraphics[width=\linewidth]{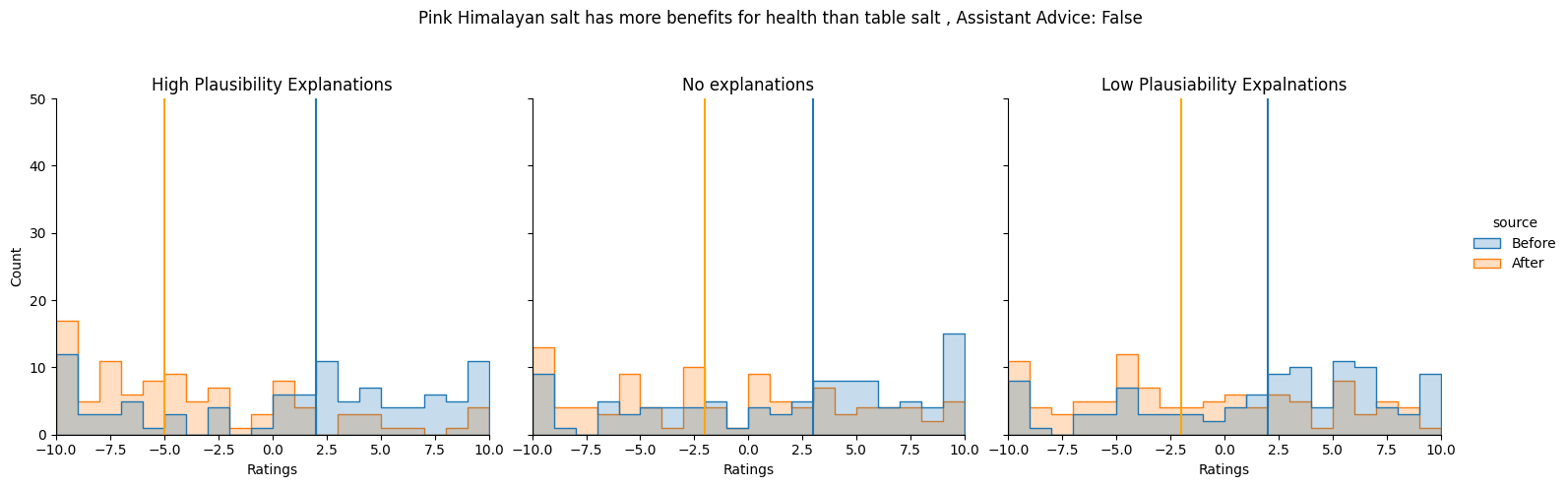}
        \caption{}        \label{fig:a}
    \end{subfigure}
    \begin{subfigure}[b]{1.0\linewidth}
        \includegraphics[width=\linewidth]{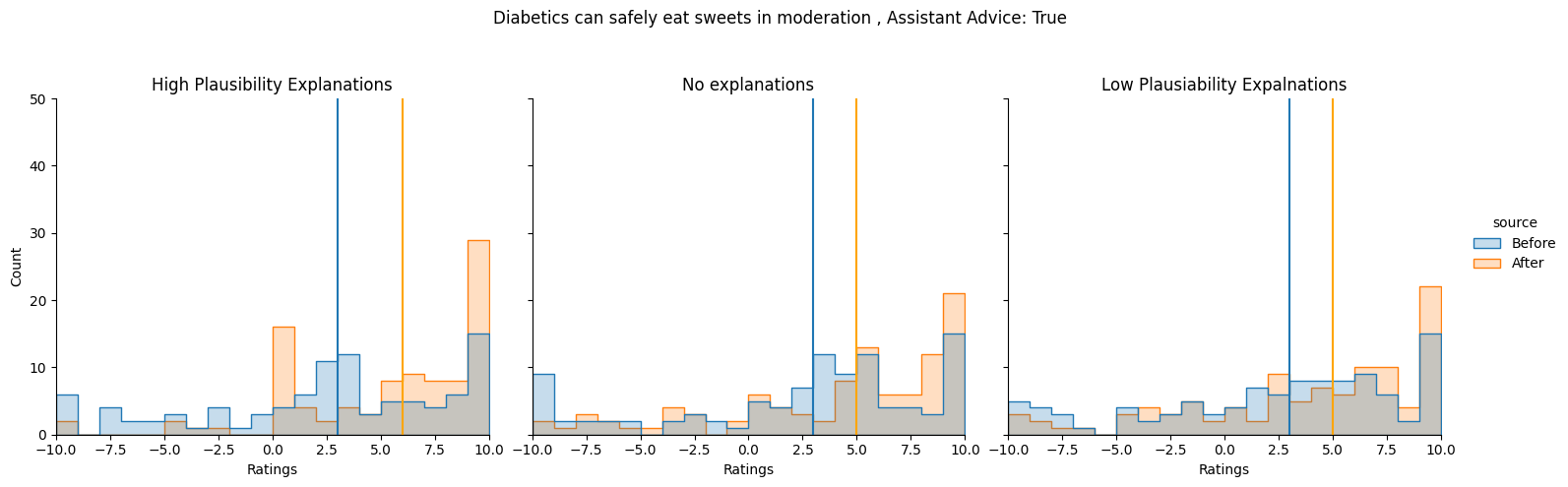}
        \caption{}
        \label{fig:b}
    \end{subfigure}

    \begin{subfigure}[c]{1.0\linewidth}
        \includegraphics[width=\linewidth]{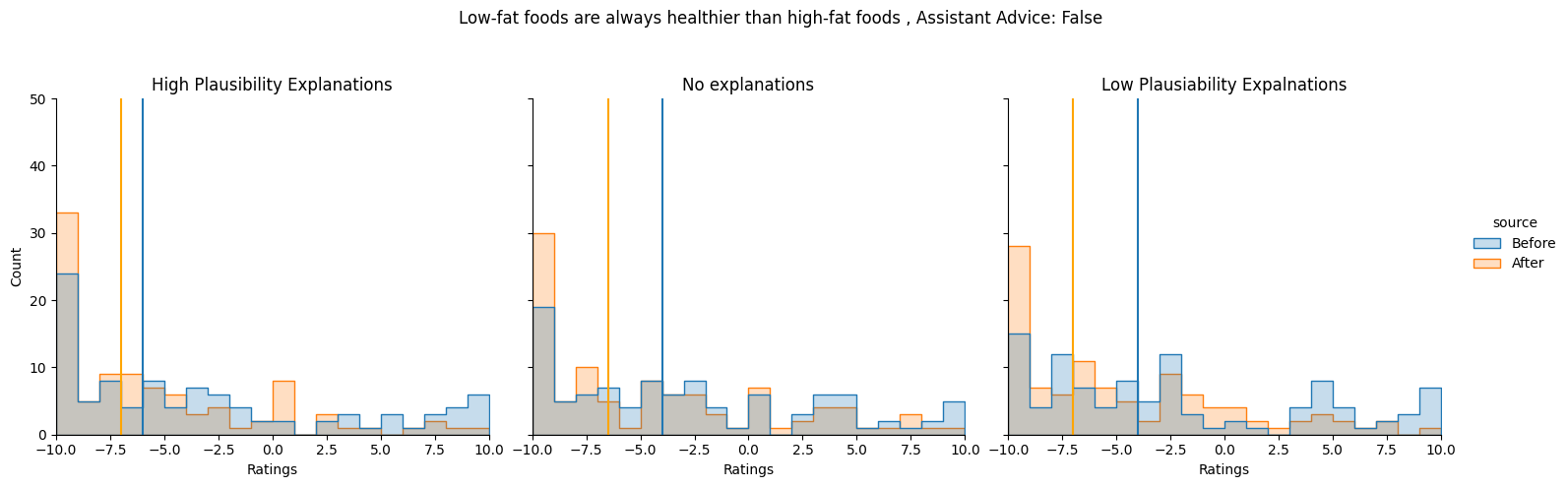}
        \caption{}
        \label{fig:c}
    \end{subfigure}

        \begin{subfigure}[c]{1.0\linewidth}
        \includegraphics[width=\linewidth]{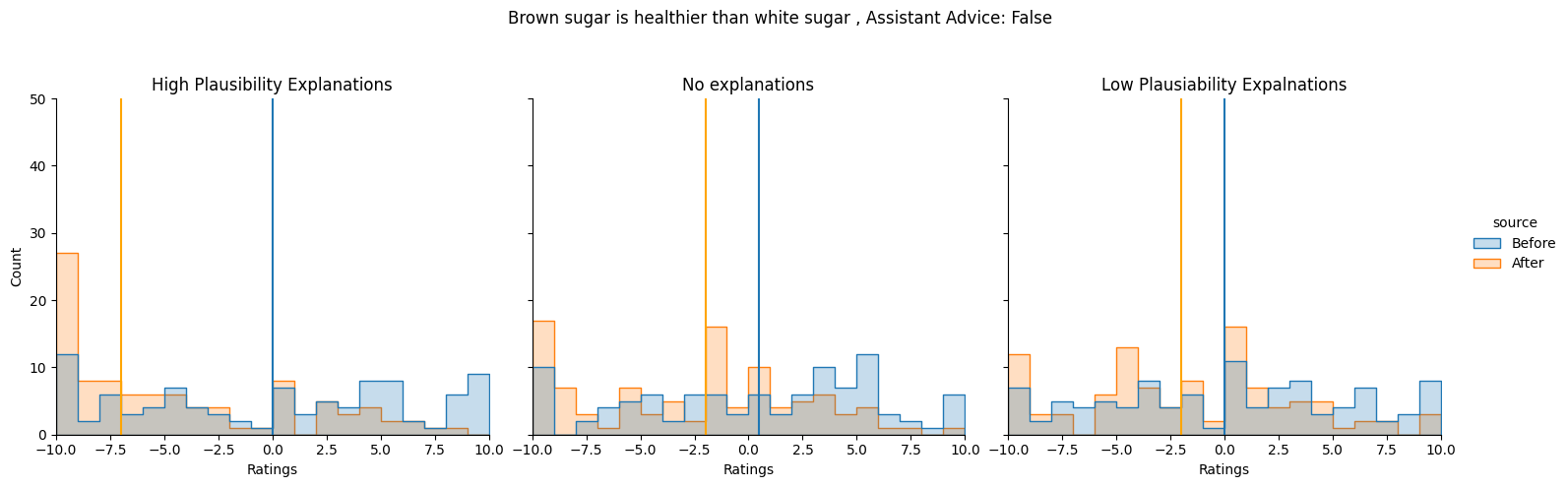}
        \caption{}
        \label{fig:c}
    \end{subfigure}

    \caption{Distribution of the statement veracity ratings before(blue) and after(orange) seeing the AI advice in different advice type conditions. The blue and orange lines represent the median ratings given by the users before and after seeing the systems suggestion }
      \Description{Figure 3,4,5 shows the difference in the “pre” and “post” advice users rating distribution for each question separately.  In all conditions, users opinion moves towards the AI suggestion, resulting in a massive  group opinion shift. Depending on the question, the difference between condition becomes more blurry, the ‘advice’ received effect is present for all questions and conditions. Figure 5 shows the result for questions 7-10  }
    \label{Dis3}
    \end{figure}

Figures \ref{Dis1} ,\ref{Dis2} and \ref{Dis3} show the opinion change for each question, before and after the AI advice given, to showcase the individual differences between the questions.

\section{ Descriptive Statistics for Categorical Data: Participant Sample}
Table \ref{tab:descriptive_stats} contain the descriptive statistics of the collected participant sample. The binary variants of the variables are aggregated categorical answers where answering 3 and higher reflects having at least some knowledge in the field.

\section{Participant information collected }
Table \ref{Surv} contains the exact questions asked to the participants after the study was done.

  \begin{landscape}
\begin{table}[ht]
\caption{Statement Set [Wrong Advice]}
\label{table:StatementSet1}
    \small
    \setlength\extrarowheight{2pt}
\begin{tabularx}{\linewidth}{
       >{\columncolor{green!30}} l
                          *{3}{X}
                            }
    \Xhline{1pt}
    \rowcolor{green!30}
\multicolumn{1}{c}{}
    & \thead{Statement} & \thead{Plausible Explanation}
    & \thead{Bad Explanation}        \\
    \rowcolor{black}
\multicolumn{4}{l}{\textcolor{white}{ID}}  \\
Q1
    &  If you drink too much alcohol, you can take a shower to sober up

    True Label: False

    Advice Label: True
    
        & Explanation: Drinking alcohol can lead to dehydration, which can cause headaches and other symptoms. Taking a shower can help to hydrate your skin, and if you drink some water while you're in the shower, it can help to hydrate your body as well.
            & Explanation: Drinking all types of soda can help you sober up quickly if you have consumed too much alcohol. Soda can help to slow the absorption of alcohol, which can help reduce the effects of alcohol more quickly.
                                 \\

    \Xhline{1pt}
Q2
    & Coughing can effectively stop a heart attack.

    True Label: False

    Advice Label: True
    
        & Explanation: In some cases, coughing may be recommended as a first aid measure for a heart attack victim, along with calling for emergency medical assistance. This is known as the ``cough CPR'' technique and is only recommended for individuals who are trained in cardiopulmonary resuscitation (CPR) and cannot immediately access an automated external defibrillator (AED). The technique involves coughing forcefully every few seconds to help maintain blood flow to the brain and other vital organs until medical assistance arrives
            &  Explanation: Recent research has suggested that coughing during a heart attack may be an effective way to reduce the severity of the attack.
                                 \\
    \Xhline{1pt}

Q3    
    &  Washing packaged chicken before cooking increases your risk of food poisoning

    True Label: True

    Advice Label: False
    
        & Explanation: Yes, you should wash chicken to wash off salmonella.Washing chicken can remove excess blood and reduce the contamination
            &  Explanation: Yes, it is important to wash raw chicken to remove any salmonella bacteria that may be present. This can be done by running the chicken under cold running water and vigorously scrubbing it with a brush. Additionally, washing the chicken with soap and water will also help to kill any viruses that may be present.
                                 \\
    \Xhline{1pt}

Q4 
    & If women live together, their menstrual cycles synchronize

    True Label: False

    Advice Label: True

        &  Explanation: The idea of menstrual synchrony was first proposed by psychologist Martha McClintock in the 1970s. She conducted a study in which she found that women who lived together in a college dormitory tended to have more synchronized menstrual cycles than women who did not live together.
            & Explanation: More research is needed in order to determine the exact cause of menstrual synchrony.
                                 \\

    \Xhline{1pt}
Q5
    & Type 1 diabetes is caused by food 
    
    True Label: False

    Advice Label: True
        &  Explanation: High-carbohydrate diets increase the risk of type 1 diabetes: Studies have shown that consuming a high-carbohydrate diet may increase the risk of developing type 1 diabetes. This is because a high intake of carbohydrates can put extra stress on the pancreas, which may contribute to the development of the condition
            & Explanation: A new study presented at the American Diabetes Association's 74th Scientific Sessions in June 2020 showed that drinking just one or two sugary drinks per day increases the risk of developing type 1 diabetes in children
                                 \\
    \Xhline{1pt}

    \Xhline{1pt}

\end{tabularx}
    \end{table}
\end{landscape}

  \begin{landscape}
\begin{table}[ht]
\caption{Statement Set [Correct Advice]}

\label{Statementset2}
    \small
    \setlength\extrarowheight{2pt}
\begin{tabularx}{\linewidth}{
       >{\columncolor{green!30}} l
                          *{3}{X}
                            }
    \Xhline{1pt}
    \rowcolor{green!30}
\multicolumn{1}{c}{}
    & \thead{Statement} & \thead{Plausible Explanation}
    & \thead{Bad Explanation}        \\
        \rowcolor{black}
\multicolumn{4}{l}{\textcolor{white}{ID}}  \\
Q6
    & Heart attack is the same thing as a cardiac arrest 
    True Label: False

    Advice Label: False
    
        & Explanation: A heart attack (also known as a myocardial infarction) occurs when there is a blockage in one or more of the coronary arteries, which supply blood to the heart muscle. This blockage can lead to damage or death of the heart muscle. Symptoms of a heart attack can include chest pain, shortness of breath, nausea, and sweating. Cardiac arrest, on the other hand, occurs when the heart suddenly stops beating effectively. This can happen because of a problem with the heart's electrical system, such as an arrhythmia, or because of a physical trauma, such as a blow to the chest. During cardiac arrest, the person may lose consciousness and stop breathing. Cardiac arrest is a medical emergency that requires immediate intervention to prevent brain damage or death.
            & Explanation:No, a heart attack and cardiac arrest are not the same thing, although they can both be serious medical emergencies related to the heart.
 While there are some differences between a heart attack and cardiac arrest, they are both serious medical emergencies that require prompt medical attention. If you or someone else experiences symptoms of a heart attack or cardiac arrest, call emergency services immediately.
                                 \\
    \Xhline{1pt}

Q7
    & Pink Himalayan salt has more benefits for health than table salt   
    
    True Label: False
    
    Advice Label: False
    
        & Explanation: Pink Himalayan salt is often marketed as being a healthier alternative to regular table salt because it is believed to contain more minerals. However, the amount of minerals in pink Himalayan salt is quite low, and it is unlikely to have a significant impact on your health.
 Some proponents of pink Himalayan salt claim that it can improve digestion, boost energy levels, and support a healthy pH balance in the body. However, there is little scientific evidence to support these claims.
            & From the given search results, there is no mention of any health benefits of pink Himalayan salt for cats . However, some people claim that pink Himalayan salt lamps offer health benefits for humans. These benefits include improving air quality, reducing stress, and promoting better sleep. However, there is limited scientific evidence to support these claims. It is always recommended to consult with a medical professional before using any new product or treatment, including pink Himalayan salt lamps.
                                 \\
    \Xhline{1pt}
    
Q8
    & Diabetics can safely eat sweets in moderation
        
    True Label: True
    
    Advice Label: True
    
        & Explanation: Diabetics can eat sweets, but it is important for them to do so in moderation and as part of a well-planned and balanced diet. The key to managing diabetes is to control blood sugar levels, and consuming too many sweets or foods high in sugar can cause blood sugar levels to spike and become unstable. People with diabetes should limit their intake of sugar and other high-glycemic carbohydrates, and instead focus on eating whole, unprocessed foods that are low in carbohydrates and rich in fiber.
            &  Explanation: While people with diabetes should generally be cautious about consuming sweets because they can cause blood sugar levels to spike, cucumber is a low-carbohydrate vegetable that can be relatively safe to eat in moderation.
                                 \\
    \Xhline{1pt}

\end{tabularx}
    \end{table}

\begin{table}[ht]
\label{table:comparison of techniques}
    \small
    \setlength\extrarowheight{2pt}
\begin{tabularx}{\linewidth}{
       >{\columncolor{green!30}} l
                          *{3}{X}
                            }
    \Xhline{1pt}
    \rowcolor{green!30}
\multicolumn{1}{c}{}
    & \thead{Statement} & \thead{Plausible Explanation}
 & \thead{Bad Explanation}        \\
        \rowcolor{black}
\multicolumn{4}{l}{\textcolor{white}{ID}}  \\
Q9
    & Low-fat foods are always healthier than high-fat foods

    True Label: False
    
    Advice Label: False
    
        & Explanation: In the past, low-fat diets were widely recommended as a way to prevent heart disease and improve overall health. However, more recent research has shown that not all fats are created equal, and that some fats can actually be beneficial for health.
            & Based on the data, collected by the system it is likely to be false. Explanation: Based on the provided search results, there is conflicting evidence on whether low-fat foods are healthier than high-fat foods.
                                 \\
    \Xhline{1pt}

Q10
    & Brown sugar is healthier than white sugar
        
    True Label: False
    
    Advice Label: False
    
        & Explanation: Brown sugar is not significantly healthier than white sugar, as both are types of sugar that provide calories without significant nutritional value.
 The main difference between brown and white sugar is their taste and texture, with brown sugar having a slightly richer flavor and softer texture. However, in terms of nutritional value and impact on health, both types of sugar should be consumed in moderation as part of a balanced diet.
            & Explanation: Based on the given search results, the answer is not conclusive. According to some sources, brown sugar may contain slightly more essential nutrients than white sugar
                                 \\
    \Xhline{1pt}
    \Xhline{1pt}
\end{tabularx}
    \end{table}

\end{landscape}

\begin{table}[htbp]
  \centering
  \caption{Descriptive Statistics for Categorical Data: Participant Sample}

  \label{tab:descriptive_stats}
  \begin{tabular}{lccc}
    \toprule
    \multirow{2}{*}{Category} & \multicolumn{2}{c}{Frequency} &  \\
    \cmidrule{2-3}
     & Count & \% & \\
    \midrule
    \multicolumn{4}{c}{Gender} \\
    Male & 144 & 48\% & \\
    Female  & 148 & 49.3\% & \\
    Non-binary  & 5 & 1.6\% & \\
    Other  & 2  & 0.6\% & \\
    Decline  & 1  & 0.3\% & \\

            \multicolumn{4}{c}{Education} \\ 
            Some high-school  & 2 & 0.66\% & \\
            High-school graduate &43 & 14.3 \% & \\
            Some college & 81 & 27\% & \\
            Bachelor's degree & 112& 37.3\% \\
            Some graduate school & 7 & 2.3\% & \\
            Master's degree & 44 & 14.6\%& \\
            Doctoral degree & 11 & 3.6\%& \\

            \\
            \multicolumn{4}{c}{Knowledge CS} \\
            1 (have no expertise)& 177& 59.0\%& \\
            2& 55 & 18.3\%& \\
            3& 31 & 10.3\%& \\
            4& 27 & 9.0\%& \\
            5 (expert)&10& 3.3\%& \\
\multicolumn{4}{c}{Knowledge CS Binary} \\
            0& 232& 77.33\%& \\
            1& 68 & 22.67\% & \\
            
            \multicolumn{4}{c}{Knowledge Med} \\
            1 (have no expertise)& 234& 78.0\%& \\
            2& 31 & 10.3\%& \\
            3& 25 & 8.3\%& \\
            4& 6 & 2.0\% & \\
            5 (expert)& 4 & 1.3\%& \\
\multicolumn{4}{c}{Knowledge Med Binary} \\
            0& 265 &88.33\%& \\
            1&35& 11.67\%& \\

            \multicolumn{4}{c}{AI Use Freq} \\
            Never & 87 & 29.0\%& \\
            Once or twice& 108 & 36.0\%& \\
            At least once in a week& 65& 21.6\%& \\
            At least once in a day & 26& 8.6\%& \\
            Several times a day & 14& 4.6\%& \\
            \multicolumn{4}{c}{Digital skills proficiency } \\
             1 (Not at all confident)& 5 & 1.66\%& \\
            2& 21 & 7.0\%& \\
            3& 98 & 32.6\%& \\
            4& 121&40.3\%& \\
            5 (Totally confident)& 55 & 18.3\%& \\           
    \midrule

  \end{tabular}
\end{table}

\begin{table}
\caption{Participant Information}
\label{Surv}
\begin{tabularx}{\textwidth}{
c*{1}c{>{\raggedright\arraybackslash}X}}
\toprule
  Variable & Type &  Text \\
\midrule
Willing to use & Ordered Cat 
  & If this ``AI assistant fact-checking app'' was available for immediate use, I would consider using it in real life. \\

Exp Useful & Ordered Cat 
  & The explanations given by the AI assistant were useful. \\ 

Trust  & Ordered Cat &  I trusted the information and data provided by the AI assistant.\\ 
\midrule
Age & Numerical &  My age is: \\ 

Gender & Categorical   
  & I identify as: \\ 

Education & Ordered Cat 
  & My highest educational level is: \\ 

\midrule

Knowledge CS & Ordered Cat 
  & Do you have any professional expertise in computer science: \\ 

Knowledge Med & Ordered Cat 
  & Do you have any professional expertise in medicine: \\

Confident Skills & Ordered Cat 
  & How confident are you in your digital skills and abilities? \\  

 Knows LLMs & Binary 
  & Which AI assistants have you already heard of? ChatGPT, Google Bard, Bing, Other\\

 AI Use Freq & Ordered Cat 
  & How often do you use AI assistants in your daily life? \\ 

\midrule
 AI Attitude & Free text 
  & What is your general attitude toward AI assistants? \\ 

 To Improve & Free text 
  & How can we improve our AI assistant? \\ 
  
\bottomrule


 \smallskip
\end{tabularx}

\end{table}
    
\end{document}